\documentclass{article}

\usepackage{PRIMEarxiv}

\usepackage[utf8]{inputenc} 
\usepackage[T1]{fontenc}    
\usepackage{hyperref}       
\usepackage{url}            
\usepackage{booktabs}       
\usepackage{amsfonts}       
\usepackage{nicefrac}       
\usepackage{microtype}      
\usepackage{lipsum}
\usepackage{fancyhdr}       
\usepackage{graphicx}       
\graphicspath{{media/}}     

\usepackage{times}
\usepackage{latexsym}

\usepackage[T1]{fontenc}

\usepackage[utf8]{inputenc}

\usepackage{microtype}

\usepackage{inconsolata}

\usepackage{graphicx}
\usepackage{subcaption}

\usepackage{algorithm}
\usepackage{multirow}
\usepackage{graphicx}
\usepackage{booktabs}
\usepackage{algorithm}
\usepackage{multicol}
\usepackage{amsmath}
\usepackage{algpseudocode}
\usepackage{caption}
\usepackage{enumitem}
\usepackage{cleveref}
\usepackage{tabularx} 
\usepackage{microtype}
\usepackage{hyperref}
\usepackage{url}
\usepackage{booktabs}
\usepackage{natbib}

\usepackage{tabularray}
\usepackage{multirow}
\usepackage{subfiles}
\usepackage{enumitem}
\usepackage{makecell}
\usepackage{graphicx}
\usepackage{wrapfig}
\usepackage[table]{xcolor} 

\usepackage{todonotes}

\definecolor{mydarkblue}{rgb}{0,0.08,0.45}
\definecolor{myblue}{HTML}{3b75c9}
\definecolor{myred}{HTML}{E33222}
\definecolor{mygreen}{HTML}{438773}
\definecolor{mymaroon}{RGB}{142,27,19}
\definecolor{maroon}{HTML}{992000}
\definecolor{mycite}{cmyk}{0.55,1,0,0.15}
\definecolor{codeblue}{rgb}{0.25,0.5,0.5}
\definecolor{codekw}{rgb}{0.85, 0.18, 0.50}
\definecolor{codegreen}{rgb}{0,0.6,0}
\definecolor{codegray}{rgb}{0.5,0.5,0.5}
\definecolor{codepurple}{rgb}{0.58,0,0.82}
\definecolor{backcolour}{rgb}{0.95,0.95,0.92}
\definecolor{refcolor}{HTML}{9F363A}
\definecolor{prompt1}{HTML}{3b75c9}
\definecolor{prompt2}{HTML}{9F363A}
\definecolor{prompt3}{rgb}{0.48,0,0.32}
\definecolor{prompt4}{HTML}{438773}
\hypersetup{
    colorlinks=true,
    citecolor=refcolor,
    linkcolor=myblue
          }
\newcommand{\User}[1]{\textbf{\textcolor{myblue}{User:}} #1}
\newcommand{\AI}[1]{\textbf{\textcolor{myred}{ChatGPT4:}} #1}
\newcommand{\athena}[1]{\href{https://athena.ohdsi.org/search-terms/start}{Athena Ontology Database}}

\title{Uncertainty Quantification for Clinical Tasks \\ with (Large) Language Models}

\author{Zizhang Chen \\
  Brandeis University \\
  \texttt{zizhang2@brandeis.edu} \\\And
  Peizhao Li \\
  Google \\
  \texttt{peizhaoli@google.com } \And
  Xiaomeng Dong \\
  GE Healthcare \\
  \texttt{xiaomeng.dong@gehealthcare.com} \And
  Pengyu Hong \\
  Brandeis University \\
  \texttt{hongpeng@brandeis.edu} \\
  }

\author{
  \vspace{1mm} Zizhang Chen$^{1}$, Peizhao Li$^{2}$, Xiaomeng Dong$^{3}$, Pengyu Hong$^{1}$ \\
  \vspace{1mm} $^{1}$Department of Computer Science, Brandeis University \\
  \vspace{1mm} $^{2}$Google\\
  \vspace{1mm} $^{3}$GE Healthcare \\
  \texttt{\{zizhang2, hongpeng\}@brandeis.edu,peizhaoli@google.com,xiaomeng.dong@gehealthcare.com} \\
}

\begin{document}
\maketitle

\begin{abstract}
To facilitate healthcare delivery, language models (LMs) have significant potential for clinical prediction tasks using electronic health records (EHRs). However, in these high-stakes applications, unreliable decisions can result in high costs due to compromised patient safety and ethical concerns, thus increasing the need for good uncertainty modeling of automated clinical predictions. To address this, we consider the uncertainty quantification of LMs for EHR tasks in white- and black-box settings. We first quantify uncertainty in white-box models, where we can access model parameters and output logits. We show that an effective reduction of model uncertainty can be achieved by using the proposed multi-tasking and ensemble methods in EHRs. Continuing with this idea, we extend our approach to black-box settings, including popular proprietary LMs such as GPT-4. We validate our framework using longitudinal clinical data from more than 6,000 patients in ten clinical prediction tasks. Results show that ensembling methods and multi-task prediction prompts reduce uncertainty across different scenarios. These findings increase the transparency of the model in white-box and black-box settings, thus advancing reliable AI healthcare. 
\end{abstract}


\section{Introduction}
\label{sec:intro}
Language models, such as~\citep{steinberg2021language, theodorou2023synthesize, steinberg2024motor} have emerged to be an efficient tool in the domain of EHR tasks. These models, extensively trained on diverse sources of clinical data, such as physician notes and longitudinal medical codes, have demonstrated remarkable effectiveness in predicting clinical outcomes. Despite their capabilities, measuring and reducing the uncertainties of these models in EHR tasks is crucial for ensuring patient safety, as clinicians can avoid interventions that the model indicates are uncertain and potentially hazardous. In addition, quantifying the uncertainties in clinical tasks can enhance the reliability of AI-driven medical decision-making systems~\citep{begoli2019need}. 

To address this challenge, leveraging the transparency of model parameters, we utilize established uncertainty metrics and propose to combine them with ensembling and multi-tasking approaches to effectively quantify and mitigate uncertainties in EHR tasks for these white-box language models. Recently, large language models have embarked on demonstrating their utility in clinical-related tasks, including EHR prediction tasks~\citep{wornow2023shaky}, analyzing radiology report examinations~\citep{jeblick2024chatgpt} and medical reasoning~\citep{lievin2024can}. However, the encapsulation of modern Large Language Models, typically offered as API services with restricted access to internal model parameters and prediction probabilities, impedes the direct application of traditional uncertainty quantification methods. To overcome this limitation, We redefine uncertainty quantification as a post-hoc approach by analyzing the distribution of answers generated repeatedly from our designed prompts for clinical prediction tasks. Inspired by the effectiveness of our proposed methods in reducing model uncertainty for white-box LMs, we adapted and applied ensembling and multi-tasking methods to the black-box settings.

The main contributions of this paper are summarized as follows:
\vspace{-3mm}
\begin{itemize}[leftmargin=*]
    \item We propose a multi-tasking method and a model ensembling approach to reduce model uncertainties for the white-box language model for clinical predictions using medical code sequences.

    \item We redefine the uncertainty quantification in EHR prediction tasks using black-box LLMs.

    \item We adapted our proposed two methods from white-box LM settings to black-box LLM settings using natural languages and demonstrated their effectiveness in reducing uncertainties.
\end{itemize} 


\section{Background and Related Work}
\label{sec:background}

\paragraph{Uncertainty Quantification in Clinical Tasks}
Uncertainty quantification has emerged as a critical component in clinical tasks, particularly in safety-critical fields such as clinical decision-making~\citep{begoli2019need, chen2021unite, tomavsev2021use} and medical imaging~\citep{edupuganti2020uncertainty, lambert2024trustworthy}. Current approaches involve applying Bayesian approaches~\citep{dusenberry2020analyzing, jahmunah2023uncertainty}, Ensembling methods~\citep{mimori2021diagnostic, abe2024pathologies} and test-time augmentations~\citep{ayhan2020expert} to reduce the model uncertainties. This work investigates the model uncertainty on structured, longitudinal EHR dataset for clinical outcome predictions.

\paragraph{Uncertainty Quantification with LLMs}

The increasing reliance on black-box large language models (LLMs) such as GPT-4~\citep{achiam2023gpt}, Claude 3~\citep{anthropic2023claude3}, and Gemini~\citep{team2023gemini} in commercial applications has introduced complex challenges in Uncertainty Quantification. Due to the closed nature of LLMs, typically offered as API services, traditional uncertainty quantification methods that require access to model parameters are not applicable. To overcome these challenges, recent research~\cite{kuhn2023semantic, lin2023generating, xiong2024can} has developed innovative techniques that estimate uncertainty based directly on the text outputs from LLMs, bypassing the need for internal data. Notably, Kuhn et al.(2023) have proposed semantic entropy as a new metric for quantifying uncertainty in LLMs, which capitalizes on the semantic equivalence across varying expressions. Subsequent studies~\citep{lin2023generating, xiong2024can} have further advanced these approaches, crafting sophisticated methods that enhance black-box UQ through strategic prompting, sampling, and result aggregation.

\begin{figure*}[t]
    \centering
    \includegraphics[width=.95\textwidth]{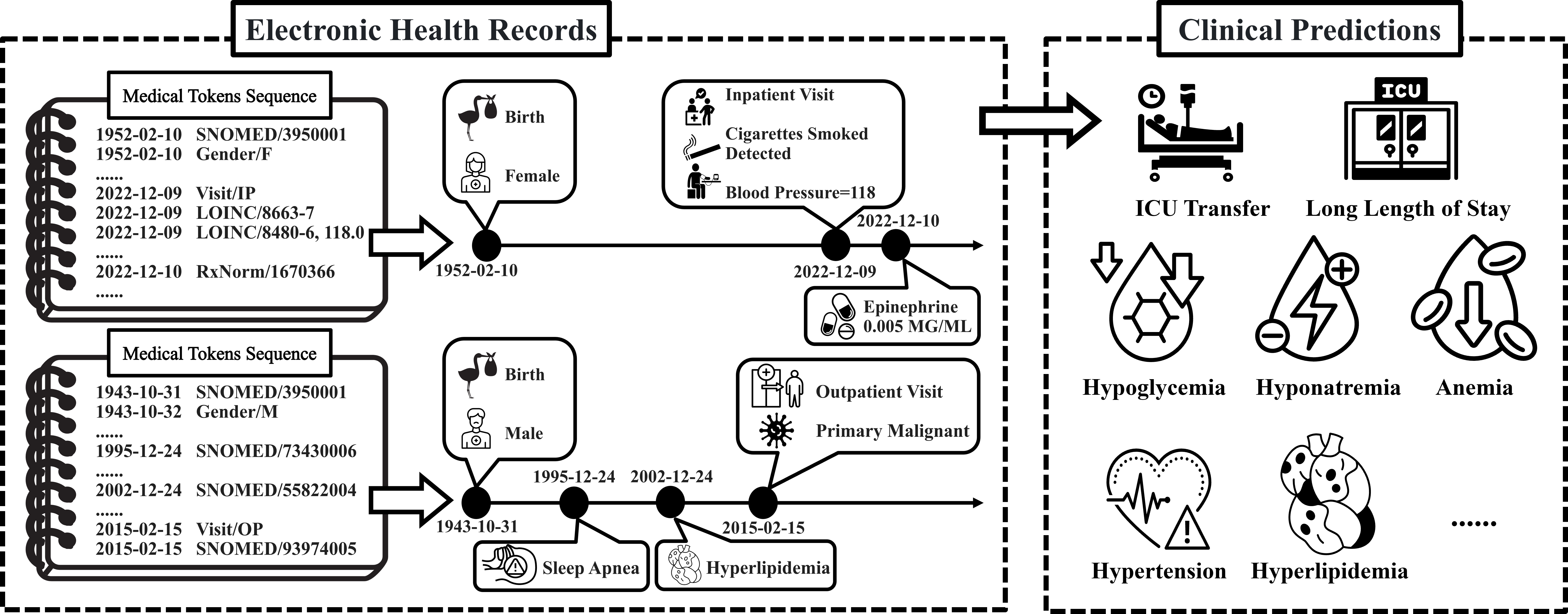}
    \caption{EHR predictions with medical codes sequences. \textbf{Left}: structured, longitudinal medical tokens, each code is in OMOP format~\citep{reich2024ohdsi} and associated with a specific time point. We translate these codes into natural languages that describe a patient's timeline. \textbf{Right}: The interpreted EHR data can be used for severing various clinical applications such as Long Length of Stay or Hypoglycemia predictions.}
    \vspace{-5mm}
    \label{fig:medseq}
\end{figure*}


\section{Predictions in Electronic Health Records}\label{sec:uq_ehr}

\paragraph{Clinical Tasks}\label{sec:tasks}
We present our findings across a range of clinical tasks using the newly published EHRSHOT dataset~\citep{wornow2023ehrshot}. The EHRSHOT dataset encompasses \textbf{structured}, \textbf{longitudinal} clinical data extracted from the electronic health records of 6,739 patients at Stanford Medicine, featuring over 40 million clinical events. Within the framework of EHRSHOT, we explore 10 EHR prediction sub-tasks organized into three distinct categories. \textit{(i)} \textit{General Operational Outcomes}: including long length-of-stay: predicting whether a patient's length of stay will exceed seven days, and ICU Transfer: predicting whether the patient will be transferred to the ICU on the same day of admission. \textit{(ii)} \textit{Lab Test results}: This task involves predicting the normalcy of lab test results immediately before their official release. The lab tests covered Thrombocytopenia, Hyperkalemia, Hypoglycemia, Hyponatremia, and Anemia. \textit{(iii)} \textit{New Diagnose Diseases}: This involves forecasting whether the patient will be first diagnosed with specific diseases within the next year from the date of discharge. Diseases tracked include Hypertension, Hyperlipidemia, and Acute Myocardial Infarction. 

\paragraph{EHR Tasks Formulation}
The EHR prediction tasks can be formulated as follows: consider patient $p$'s data represented as $\{C_{p}, Y_{p}\}$. Here, medical sequence $C_{p}$$=$$\{(c_{1}, t_{1}), (c_{2}, t_{2}), ..., (c_{n}, t_{n})\}.$ denotes the complete clinical events for patient $p$ arranged chronologically. Each code $c_{i}$ represents a specific medical event at time point $t_{i}$ in Observational Medical Outcomes Partnership (OMOP) format~\citep{reich2024ohdsi}. $Y_{p}$$=$$\{y_{1}, y_{2}, ... y_{m}\}_{m\in\{t_1, t_2, ...t_{n}\}}$ is a set of labels indicating the clinical outcomes. At time $t_k$, our objective is to predict label $y_{t_k}$, with a truncated medical sequence Here, we denote $C_{t_{k-1}}$$=$$\{(c_{1}, t_{1}), (c_{2}, t_{2}), ...(c_{k-1}, t_{k-1})\}$, which consists of data up to, but not include time point $t_k$.



\section{Modeling Uncertainties for Clinical Outcome Predictions in EHR}



We consider two settings for uncertainty modeling: white-box and black-box. In white-box settings, the access to model's parameters and output probabilities/logits is available. We employ four widely recognized metrics: Brier Score~\citep{rufibach2010use}, Expected Calibration Error~\citep{naeini2015obtaining}, Adaptive Expected Calibration Error~\citep{nixon2019measuring}, and Negative Log Likelihood. We also employ two additional methods, Deep Ensemble and Monte Carlo Dropout, to reduce the model uncertainties for clinical predictions. We detail the formulation of the uncertainty metrics and implementing uncertainty methods in~\cref{sec:model_reduce_white}. Reducing the model uncertainties can enhance the model's trustworthiness, especially in clinical decision-making. In the black-box setting, where the intrinsic model parameters are unavailable, one approach to quantify uncertainty is calculating entropy-based metrics on a repeatedly generated answer set. Then, these metrics are used to predict whether to rely on the model's response or not~\citep{filos2019benchmarking, kuhn2023semantic}. We utilize uncertainty metrics to quantify the trustworthiness of the proprietary model in EHR tasks.


\subsection{Modeling and Reducing Uncertainties in White-box Settings}
\label{sec:model_reduce_white}
\paragraph{Clinical Prediction with BERT-Based Language Models}
In our \textit{white box} setting for EHR predictions, we first follow the settings of~\citep{steinberg2021language, wornow2023ehrshot} to generate sequence embeddings $\{e_{i}\}_{i=1}^{k-1}$ from medical code sequence $C_{t_{k-1}}$ using the CLMBR-T-base, a foundation model pre-trained on 2.57 million deidentified structured patient records from the private Stanford HealthCare Data Warehouse. The CLMBR-T-base model is pre-trained autoregressively to predict the subsequent medical code based on their prior medical sequence. The published CLMBR-T-base model includes encoders that generate representations of a patient's medical tokens. It contains embedding layers that first map up to $65536$ unique medical tokens into a hidden dimension space of $768$, then followed by a $12$ stacked transformer layer with a fixed context window of $496$ tokens. Multiple token representations could exist at the time point $t_{k-1}$; however, the last possible representation $e_{k-1}$ is used for the downstream EHR predictions. According to~\cite {wornow2023ehrshot}, this selection strategy not only maximizes the utilization of available information but also helps prevent data leakage by ensuring that future data do not influence prediction. We subsequently trained decoders using the embedding $e_{{k-1}}$ for each downstream EHR prediction task as introduced in~\cref{sec:tasks}. We then assess model uncertainties and implement the uncertainty methods across decoders.

\paragraph{Uncertainty Metrics}
We adopt Brier Score, Expected Calibration Error, Adaptive Expected Calibration Error, and Negative Log Likelihood to quantify and assess the model uncertainties in EHR prediction tasks within our \textit{white-box} settings. 
We explain the formulations for these metrics: 
Brier Score \textbf{(BS)} is formulated to evaluate the accuracy of model's probabilistic predictions:
\begin{align}
    \text{BS} &= \frac{1}{N} \sum_{i=1}^{N} (y_i - \hat{p}_i)^2. \label{eq:BS}
\end{align}
where $y_i$ represents the actual label and $\hat{p}_i$ is the predicted probability of a clinical outcome for each case $i$. A lower Brier score reflects the higher confidence of the model in its classification predictions. 

Excepted Calibration Error/Adaptive Excepted Calibration Error \textbf{(ECE/aECE)} is used to measure the calibration error of the classification models: 
\begin{align}
    \text{ECE} &= \sum_{m=1}^{M} \frac{|B_m|}{N} \left| \text{acc}(B_m) - \text{conf}(B_m) \right| \label{eq:AECE}. 
\end{align}
which calculate the difference between predicted probabilities and actual outcomes in $M$ bins, which are formed by dividing predicted probabilities into a series of intervals. Here, $B_m$ are prediction bins and $\text{acc}(B_m)$ and $\text{conf}(B_m)$ represent the precision and the average predicted probability within each bin, respectively. ECE uses fixed-width bins, while aECE adjusts the bin widths based on the data distribution. A lower ECE/aECE suggests that a probabilistic model is well-calibrated, indicating a close correspondence between the predicted probabilities and the actual clinical outcomes. 

Negative log-likelihood \textbf{(NLL)} measures the probability of the actual data given the model parameters, and representation set $x_i$: 
\begin{align}
    \text{NLL} &= -\sum_{i=1}^{N} \log(p(y_i | x_i)). \label{eq:NLL}
\end{align}
A lower Negative log-likelihood indicates that the model assigns high probabilities to the correct outcomes, implying high confidence in its predictions.
aECE/ECE directly measures the model calibration error but does not penalize the model for being unconfident. The BS measures the confidence level for the prediction, and NLL significantly penalizes the model for being overconfident about false predictions. Integrating the above three metrics measures the overall model uncertainties. 

\paragraph{Uncertainty Methods}
We implement \textit{two uncertainty methods} and propose \textit{one framework} to quantify and reduce model uncertainty in the decoders used for clinical predictions. We introduce two uncertainty methods: Monte Carlo Dropout~\citep{gal2016dropout} and Deep Ensemble~\citep{lakshminarayanan2017simple, rahaman2021uncertainty}. Monte Carlo Dropout applies dropout not only during the training of neural networks but also during inference stages. This approach approximates Bayesian inference in deep Gaussian processes and allows the model to generate a distribution of predictions. Thus, it quantifies the uncertainty by allowing the network to express its confidence level through the variability of its predictions under different neuron configurations. Deep ensembles are created by training multiple versions of the same decoders, with variations only in random seeds and hyperparameters. The predictions of individual models are aggregated to produce a final prediction. This approach mitigates the inherent bias on the models, thereby reducing the overall uncertainty of the model. In addition, we propose a simple yet effective multitasking framework to predict multiple clinical tasks within the same category (\textit{General Operation Outcome}, \textit{Lab Test Results}, and \textit{New Diagnose Diseases}) presented simultaneously in~\cref{sec:tasks}. For simplicity, we formulate the multi-tasking framework as follows:
\begin{align}
\label{eq:multi}
y_{t_{k-1}}^{h_i} &= f(e_{k-1} \oplus e_{h_i}).
\end{align}
Here, $e_{k-1}$ is the representation embeddings for the clinical sequences $C_{t_{k-1}}$, $e_{h_i}$ denotes the task-specific embeddings that are combined with $e_{k-1}$ to distinguish among different clinical subtasks, $h_i$ refers to a specific clinical task, and $H$ represents the clinical task category within the same category. 

We present our results~\cref{tab:white-box}. Observing Deep Ensemble's multitasking approach reduces uncertainties with white-box models by a large margin in EHR tasks. We are compelled to ask: 
\begin{center}
\fontsize{10.25pt}{10.5pt}\selectfont
\textit{Can uncertainty reduction apply to proprietary LLMs?} 
\end{center}
This propels us to investigate specialized uncertainty quantification for the proprietary \textbf{black-box} models like GPTs in EHR prediction tasks.


\subsection{Transferring to Black-box Setting}


\paragraph{Clinical Prediction with GPTs}
In our \textit{black-box} settings where the model parameters are opaque, we transform the patient's medical code sequence $C_{t_{k-1}}$ into free-form languages by generating text descriptions for each medical token. This process leverages information from the \href{https://athena.ohdsi.org/search-terms/start}{Athena Ontology Database}~\citep{hripcsak2015observational, reich2024ohdsi}, where the concept of each medical code is clearly defined, categorized and described. We then adapted the descriptions by retaining only the most recent medical code details to fit within the input context length constraints of the GPT models. We use $S_{t_{k-1}}$ to denote the text descriptions of the medical sequences of a patient. We construct prompt $P_{t_{k-1}}$$=$$\{G, S_{t_{k-1}}, \Omega, O\}$ based on $S_{t_{k-1}}$, where $G$ denotes general prompts which specify the role and scenario LLM acts, $\Omega$ specifies the clinical tasks to be performed, $O$ defines the output formats of the GPT's response. We instruct the LLM to repeatedly generate a set of $n$ responses $R$$=$$\{r_1, r_2, ...,r_n\}$ from prompt $P_{t_{k-1}}$. We then conduct post-hoc uncertainty quantification over the response set $R$. In the following sections, we detail the methods for quantifying and reducing the uncertainties in white-box LM settings and proprietary LLM settings for EHR prediction tasks. 

\paragraph{Uncertainty Quantification for EHR tasks with GPTs}
We introduce the uncertainty quantification method and develop the corresponding metrics explicitly designed for proprietary large-language models, which are applied in the context of clinical prediction tasks. Similarly to the existing pipeline on proprietary LLM uncertainty quantification~\citep{kuhn2023semantic, liu2024uncertainty}, we characterize black-box uncertainty quantification methods as post hoc approaches, required by the limited access to internal model parameters.

\begin{enumerate}
\item For a clinical prediction task on an EHR for the patient $p$, we generate a patient text description $S_p$ from its medical sequence $C_p$.\vspace{-2mm}
\item Construct the prompt $P$$=$$\{G, S_{p}, \Omega, O\}$ by integrating the text description with the role and scenario prompt $G$, the clinical task prompt $\Omega$, and the output format prompt $O$.\vspace{-2mm}
\item For clinical tasks with a specific prediction time $t_k$, we customized the prompt $P$ accordingly: first, all clinical events after the time point $t_{k}$ are excluded to prevent data leakage; second, we limit the number of clinical events included to fit within the context length constraints of the Large Language Model. We denote the tailored clinical events descriptions as $S_{t_{k-1}}$ and the customized prompt as $P_{t_{k-1}}$\vspace{-2mm}
\item Instruct Large Langugage Models to generate $n$ responses for the prompt $P_{t_{k-1}}$. Consequently, we obtain a set of responses $R_{t_{k-1}} = \{r_{t_{k-1}}^{1}, r_{t_{k-1}}^{2}, ..., r_{t_{k-1}}^{n}\}$. Here, $r_{t_{k-1}}^{i}$ represents the outputs of $i^{th}$ response.\vspace{-2mm}
\item Calculate the uncertainty score $U$ from the response repeatedly generated $R_{t_{k-1}}$.\vspace{-2mm}
\end{enumerate}
\textit{In Step 1}, we expand each medical token with the corresponding time point $(c_i, t_i)$$\in$$C_p$ into free-formed languages that describe medical events at time $t_i$. This is mainly because the GPT does not fully understand the medical codes used in various formats across hospitals and organizations use~\citep{soroush2024large}, while it can somewhat understand the free-formed clinical languages. Here, we present a short dialogue question between the user and GPT for demonstrations in~\cref{tab:prompt1}. The first prompt is a direct ask for the OMOP code. The second prompt converts the OMOP codes into clinical descriptions using the \href{https://athena.ohdsi.org/search-terms/start}{Athena Ontology Database}. 

\begin{table}[t]
    \centering
    \label{tab:prompt1}
        \small
        \begin{tabular}{|>{\raggedright\arraybackslash}p{7.25cm}|}
        \midrule
        \User{What is code "CVX/20?"} \vspace{1mm}\\
        \AI{}\\
        Answer 1: The term "CVX/20" doesn't refer to a widely recognized standard or concept ... \vspace{1mm}\\
        Answer 2: The code "CVX/20" typically refers to a vaccine code in the context of medical records ... However, I'd need more context to provide a specific answer about this code ... \vspace{1mm} \\
        \midrule
        \User{What is: "diphtheria, tetanus toxoids and acellular pertussis vaccine," can you briefly explain?
                } \vspace{1mm}\\
        \AI{}\\
        Answer: The "Diphtheria, Tetanus Toxoids and Acellular Pertussis Vaccine," commonly abbreviated as DTaP, is a combination vaccine that protects against three serious diseases caused by bacteria  ...  This vaccine is crucial for preventing these diseases in children, who are particularly susceptible to serious complications from them ... \vspace{1mm}\\
        \midrule
        \end{tabular}
        \caption{A card of a sample dialogue between a user and ChatGPT4 regarding medical codes and their descriptions: GPTs have a limited understanding of OMOP medical tokens; however, they can relatively grasp the clinical meanings denoted by these tokens.}
\end{table}

\textit{For steps 2 and 3}, we constructed our prompts into four parts: \textcolor{prompt1}{\textit{(i)}} A role-playing prompt $G$ with task-specific instructions; here, we instruct GPT to act as experienced doctors capable of providing clinical insights. \textcolor{prompt2}{\textit{(ii)}} Clinical descriptions $S_p$ that indicate the patient $p$'s medical events. We begin by extracting the patient's demographic information and calculating the patient's age at the prediction time from $C_p$ using the \href{https://www.ohdsi.org/data-standardization/}{OMOP Common Data Model}~\citep{hripcsak2016characterizing, reich2024ohdsi}. \textcolor{prompt3}{\textit{(iii)}} Questions and Clinical Tasks $\Omega$ to be answered. We develop task-specific prompts for each of the ten clinical tasks within the EHRSHOT dataset. \textcolor{prompt4}{\textit{(iv)}} Output format prompt $O$ that requests GPT to provide restricted responses for clinical tasks (E.g., "Yes, the patient will be transferred to ICU." or "No." otherwise. For predicting whether the patient will be transferred to ICU on the day on admission). We present~\cref{tab:prompt2} to demonstrate our prompt design for guiding GPTs in EHR tasks. Due to the lengthy context of clinical descriptions, we repeat our task questions at both the beginning and ending of a prompt, utilizing the Needle-In-A-Haystack (NIAH) method with GPT models. \textit{For steps 4 and 5}, given a set of repeatedly generated responses:  $R_{t_{k-1}} = \{r_{t_{k-1}}^{1}, r_{t_{k-1}}^{2}, ..., r_{t_{k-1}}^{n}\}$. We adopt the methodology outlined in~\citep{kuhn2023semantic} to compute Class Entropy $U({R_{t_{k-1}}})$ as \textit{Uncertainty Score}. $U({R_{t_{k-1}}})$'s formula presented as: 
\begin{equation}
\label{eq:ce}
    U({R_{t_{k-1}}}) = -\sum_{a_i \in A} P(a_i) \log P(a_i),
\end{equation}
where $a_i$$\in$$A$ is the clinical outcome label predicted by GPTs, the probability $P(a_i)$ of each clinical class $i$ is determined by the frequency of occurrences of class $a_i$ within the answer class set $A$. Following~\cite{kuhn2023semantic, lin2023generating}, we construct our Uncertainty Metric using the Uncertainty Score $U({R_{t_{k-1}}})$ to predict whether LLM can correctly generate an answer. We employ the area under the receiver operating characteristic curve (AUROC) as the metric for assessing uncertainty.

\begin{table}[t]
    \centering
    \label{tab:prompt2}
        \small
        \begin{tabular}{|>{\raggedright\arraybackslash}p{7.25cm}|}
        \midrule
        \textcolor{prompt1}{Role:} Assuming you are an experienced doctor. Based on the descriptions of the patient's age, demographics, and medical events provided, Use your knowledge and reasoning to predict whether $\{\textit{}{Tasks}\}$. \\ 
        \textcolor{prompt1}{Chain of thoughts:} \\
        1. Review Patient Profile: Analyzing age, sex, and medical history ... \\
        2. Evaluate Current Symptoms: Identify vital signs outside the normal range ... \\
        ......\\
        \midrule
        \textcolor{prompt2}{Patient age and demographic information:} \vspace{1mm}\\
        The patient was 36 years old at the discharge time. The patient has the following demographic information: ... \\
         \textcolor{prompt2}{Medical Events:} \\
        \textit{On June 12, 2014:} \\
        One clinical drug event, "Oxycodone hydrochloride 5 MG Oral Tablet" was recorded. \\
        One clinical Drug event, "Acetaminophen 10 MG/ML Injectable Solution" recorded. \\
        \textit{On June 13, 2014:} \\
        One clinical drug event, "Oxycodone hydrochloride 5 MG Oral Tablet" was recorded. \\
        Four measurement events, "Systolic blood pressure" was recorded with values: 131.0, 127.0, 133.0, 143.0. \\
        ......\\
        \midrule
        \textcolor{prompt3}{Tasks:}
        $\{\textit{}{Tasks}\}$.
        \vspace{1mm}\\
        \midrule
        \textcolor{prompt4}{Output format:} \\ Please answer with "Yes" or "No". \vspace{1mm}\\
        \midrule
        \end{tabular}
        \caption{An example card of prompts for EHR tasks, structured into four parts. The first section comprises a general prompt incorporating role-playing and chain-of-thought reasoning using GPTs. The second section comprises clinical descriptions in natural languages converted from medical codes. The third and fourth sections describe the clinical task and output restrictions.}
\end{table}


\begin{table*}[ht]
\centering
    \vspace{-3mm}
    
    \resizebox{1
    \columnwidth}{!}{
    \begin{tabular}{lcccc|cccc|cccc}
    \toprule
    &\multicolumn{4}{c|}{\textbf{Single-tasking Baseline}} &\multicolumn{4}{c|}{\textbf{Single-tasking Deep Ensemble}} &\multicolumn{4}{c}{\textbf{Single-tasking MC dropout}} \\  
    \midrule 
    \textbf{EHR Task / U.Q Metric} & $Brier\downarrow$ & $NLL\downarrow$ & {$ECE\downarrow$} & {$aECE\downarrow$}
    & $Brier\downarrow$ & $NLL\downarrow$ & {$ECE\downarrow$} & {$aECE\downarrow$}
    & $Brier\downarrow$ & $NLL\downarrow$ & {$ECE\downarrow$} & {$aECE\downarrow$}
    \\
    Long Length of Stay & \cellcolor{purple!6}0.5645 & \cellcolor{purple!6}1.5253 & \cellcolor{purple!6}0.2575 & \cellcolor{purple!6}0.2575 & \cellcolor{purple!6}0.4942 & \cellcolor{purple!6}0.8746 & \cellcolor{purple!6}0.1661 & \cellcolor{purple!6}0.1696 & \cellcolor{purple!6}0.5416 & 1.3656 & \cellcolor{purple!6}0.2232 & \cellcolor{purple!6}0.2265 \\
    ICU Transfer & 0.1388 & \cellcolor{purple!6}0.4266 & 0.0530 & 0.0527 & 0.1013 & \cellcolor{purple!6}0.3291 & 0.0425 & 0.0411 & 0.1108 & \cellcolor{purple!6}0.4432 & 0.0500 & 0.0424 \\
    \midrule
    \vspace{-4.5mm}
    \\
    Thrombocytopenia & 0.5002 & 0.6935 & 0.4861 & 0.4857 & 0.0395 & 0.1364 & 0.0183 & 0.0219 & 0.5004 & 0.6938 & 0.4858 & 0.4849 \\
    Hyperkalemia & 0.5006 & 0.6939 & 0.4809 & 0.4796 & 0.0496 & 0.1772 & 0.0252 & 0.0272 & 0.5002 & 0.6933 & 0.4806 & 0.4801 \\
    Hypoglycemia & 0.6345 & 1.4092 & 0.2727 & 0.2706 & 0.5160 & 0.7986 & 0.1357 & 0.1357 & 0.6178 & 1.1984 & 0.2399 & 0.2400 \\
    Hyponatremia & 0.7079 & 1.6924 & 0.3162 & 0.3162 & 0.5613 & 0.9057 & 0.1748 & 0.1748 & 0.6701 & 1.3312 & 0.2745 & 0.2745 \\
    Anemia & 0.7212 & 1.7921 & 0.3213 & 0.3211 & 0.5888 & 0.9668 & 0.1986 & 0.1986 & 0.6673 & 1.4020 & 0.2750 & 0.2737 \\
    \midrule
    \vspace{-4.5mm}
    \\
    Hypertension & \cellcolor{purple!6}0.2763 & \cellcolor{purple!6}0.8392 & \cellcolor{purple!6}0.1149 & \cellcolor{purple!6}0.1134 & 0.2587 & \cellcolor{purple!6}0.5458 & 0.0905 & 0.0893 & 0.2756 & \cellcolor{purple!6}0.7987 & 0.1181 & 0.1101 \\
    Hyperlipidemia & \cellcolor{purple!6}0.2495 & \cellcolor{purple!6}0.8229 & 0.1157 & \cellcolor{purple!6}0.1111 & \cellcolor{purple!6}0.2187 & \cellcolor{purple!6}0.5106 & \cellcolor{purple!6}0.0825 & \cellcolor{purple!6}0.0810 & 0.2453 & \cellcolor{purple!6}0.7004 & 0.1051 & 0.0967 \\
    Acute MI & 0.0769 & \cellcolor{purple!6}0.2480 & 0.0315 & 0.0282 & 0.0690 & \cellcolor{purple!6}0.1676 & 0.0302 & \cellcolor{purple!6}0.0226 & 0.0900 & \cellcolor{purple!6}0.2308 & 0.0339 & \cellcolor{purple!6}0.0268 \\
    \midrule
    \vspace{-2.5mm}
    \\
    \toprule
    &\multicolumn{4}{c|}{\textbf{Multi-tasking Baseline}} &\multicolumn{4}{c|}{\textbf{Multi-tasking Ensemble}} &\multicolumn{4}{c}{\textbf{Multi-tasking MC dropout}} \\ 
    \midrule
    \textbf{EHR Task / U.Q Metric} & $Brier\downarrow$ & $NLL\downarrow$ & {$ECE\downarrow$} & {$aECE\downarrow$}
    & $Brier\downarrow$ & $NLL\downarrow$ & {$ECE\downarrow$} & {$aECE\downarrow$}
    & $Brier\downarrow$ & $NLL\downarrow$ & {$ECE\downarrow$} & {$aECE\downarrow$}
    \\
    Long Length of Stay  & 0.6215 & 1.5723 & 0.2798 & 0.2789 & 0.5257 & 0.9421 & 0.1890 & 0.1896 & 0.5703 & \cellcolor{purple!6}1.3140 & 0.2381 & 0.2381 \\
    ICU Transfer  & \cellcolor{purple!6}0.0993 & 0.7288 & \cellcolor{purple!6}0.0477 & \cellcolor{purple!6}0.0466 & \cellcolor{purple!6}0.0922 & 0.3905 & \cellcolor{purple!6}0.0399 & \cellcolor{purple!6}0.0382 & \cellcolor{purple!6}0.0969 & 0.6399 & \cellcolor{purple!6}0.0455 & \cellcolor{purple!6}0.0418 \\
    \midrule
    \vspace{-4.5mm}
    \\
    Thrombocytopenia & \cellcolor{purple!6}0.0344 & \cellcolor{purple!6}0.1532 & \cellcolor{purple!6}0.0199 & \cellcolor{purple!6}0.0227 & \cellcolor{purple!6}0.0301 & \cellcolor{purple!6}0.0947 & \cellcolor{purple!6}0.0108 & \cellcolor{purple!6}0.0203 & \cellcolor{purple!6}0.0369 & \cellcolor{purple!6}0.1293 & \cellcolor{purple!6}0.0231 & \cellcolor{purple!6}0.0300 \\
    Hyperkalemia & \cellcolor{purple!6}0.0459 & \cellcolor{purple!6}0.1692 & \cellcolor{purple!6}0.0246 & \cellcolor{purple!6}0.0284 & \cellcolor{purple!6}0.0408 & \cellcolor{purple!6}0.1197 & \cellcolor{purple!6}0.0155 & \cellcolor{purple!6}0.0234 & \cellcolor{purple!6}0.0474 & \cellcolor{purple!6}0.1490 & \cellcolor{purple!6}0.0272 & \cellcolor{purple!6}0.0325 \\
    Hypoglycemia  & \cellcolor{purple!6}0.5001 & \cellcolor{purple!6}0.6945 & \cellcolor{purple!6}0.2230 & \cellcolor{purple!6}0.2208 & \cellcolor{purple!6}0.5020 & \cellcolor{purple!6}0.6972 & \cellcolor{purple!6}0.0787 & \cellcolor{purple!6}0.0882 & \cellcolor{purple!6}0.4959 & \cellcolor{purple!6}0.6929 & \cellcolor{purple!6}0.2119 & \cellcolor{purple!6}0.2067 \\
    Hyponatremia  & \cellcolor{purple!6}0.5649 & \cellcolor{purple!6}1.1548 & \cellcolor{purple!6}0.2236 & \cellcolor{purple!6}0.2225 & \cellcolor{purple!6}0.5047 & \cellcolor{purple!6}0.8144 & \cellcolor{purple!6}0.1702 & \cellcolor{purple!6}0.1701 & \cellcolor{purple!6}0.5242 & \cellcolor{purple!6}0.9476 & \cellcolor{purple!6}0.1767 & \cellcolor{purple!6}0.1757 \\
    Anemia  & \cellcolor{purple!6}0.6797 & \cellcolor{purple!6}1.3099 & \cellcolor{purple!6}0.2856 & \cellcolor{purple!6}0.2824 & \cellcolor{purple!6}0.5605 & \cellcolor{purple!6}0.8118 & \cellcolor{purple!6}0.1586 & \cellcolor{purple!6}0.1586 & \cellcolor{purple!6}0.6386 & \cellcolor{purple!6}1.0853 & \cellcolor{purple!6}0.2379 & \cellcolor{purple!6}0.2366 \\
    \midrule
    \vspace{-4.5mm}
    \\
    Hypertension  & 0.2778 & 0.9824 & 0.1213 & 0.1205 & \cellcolor{purple!6}0.2549 & 0.5810 & \cellcolor{purple!6}0.0903 & \cellcolor{purple!6}0.0793 & \cellcolor{purple!6}0.2601 & 0.8215 & \cellcolor{purple!6}0.1075 & \cellcolor{purple!6}0.1019 \\
    Hyperlipidemia & 0.2503 & 1.0269 & \cellcolor{purple!6}0.1151 & 0.1147 & 0.2410 & 0.6209 & 0.0936 & 0.0927 & \cellcolor{purple!6}0.2283 & 0.9220 & \cellcolor{purple!6}0.1023 & \cellcolor{purple!6}0.0944 \\
    Acute MI & \cellcolor{purple!6}0.0636 & 0.3306 & \cellcolor{purple!6}0.0299 & \cellcolor{purple!6}0.0258 & \cellcolor{purple!6}0.0615 & 0.1892 & \cellcolor{purple!6}0.0276 & 0.0252 & \cellcolor{purple!6}0.0675 & 0.3405 & \cellcolor{purple!6}0.0292 & 0.0289 \\
    \bottomrule
    \end{tabular}
    }\vspace{-1mm}
    \caption{ \small
    Uncertainty Metrics for clinical predictions with BERT-Based Language Models. \textbf{Top}: Uncertainty metrics for single task settings. Ten decoders are trained for each clinical baseline task. \textbf{Bottom}: Uncertainty metrics for multi-task settings. Three decoders are trained for each baseline clinical task category. \textbf{left}: Baseline, \textbf{Middle}: Deep Ensembles, \textbf{Bottom}: MC Dropout.
    }
    \vspace{-3mm}
    \label{tab:white-box}
\end{table*}

\subsection{Reducing Uncertainties for proprietary black-box models}
\label{sec:reduce_uq}
In this section, we employ two approaches to reduce the uncertainty in EHR predictions generated by GPTs. Our first approach involves ensembling clinical predictions from multiple GPT models. For a specific prompt generated from a patient's sequence of medical codes, we repeatedly generate response sets from GPT-3.5-Turbo and GPT-4. We then combine two sets of GPT responses and compute the uncertainty score $U$. This idea is drawn from the proven efficacy of ensemble methods in reducing uncertainty for white-box deep learning models, as substantiated by key studies~\citep{lakshminarayanan2017simple, rahaman2021uncertainty, abe2023pathologies} and our empirical findings in~\cref{tab:white-box} with clinical EHR tasks. Similarly, based on our empirical findings from~\cref{exp:black-box}, where predicting multiple clinical tasks within the same category can marginally reduce the uncertainty of the white box model. We extend this methodology to large language models of the GPT style. Our second approach involves instructing GPT models to generate predictions for several EHR tasks within the same clinical task category in a single-generation process. Similarly to~\cref{eq:multi}, we formulate the multi-task framework for GPTs as follows:
\begin{align}
\label{eq:mult_uq}
\{y^{h_i}\}_{h_{i}\in H} &= LLM(\{G, S, \Omega, O\}).
\end{align}
Here, $h_i$ refers to a specific clinical task, and $H$ represents the set of all clinical tasks within the same category in~\cref{sec:tasks}. $\{G, S, \Omega, O\}$ is our constructed prompt for the clinical tasks $S$. ${\{y^{h_i}\}}$ represents the ensemble of EHR predictions we prompt the LLM to generate simultaneously. 

\section{Experiment}
\subsection{Data Setup}
Following~\citep{wornow2023ehrshot}, we use the EHRSHOT benchmark to set up our experiments on structured longitudinal EHR datasets. We selected ten clinical tasks to evaluate the model uncertainty of the white-box model in clinical predictions. In the data preparation phase, we tailor the EHRSHOT dataset for each sub-task within a given task category. This approach ensures that, for each prediction category, every patient's medical sequence is associated with multiple corresponding labels at the same prediction time point. To assess the uncertainty of proprietary GPTs, we stochastically choose 100 medical sequences from each category of clinical tasks in the EHRSHOT database. The selection method ensures that each dataset includes a sufficient number of positive labels. We present our dataset information in Table~\ref{tab:dataset_info}.

\begin{table}[h]
\centering
\resizebox{0.5
\columnwidth}{!}{
\begin{tabular}{lccc}
    \toprule
    White-box Data & \#Patient & \#Events & \# Train/Val/Test\\
    \midrule
    \textit{Operational Outcome} & 3,617 & 6,491 & 2,402 / 2,052 / 2,037 \\
    \textit{Lab Tests} & 5,691 & 152,331 & 59,983 / 44,928 / 47,420 \\
    \textit{New Diagnose} & 1,916 & 2,794 & 959 / 956 / 879 \\
    \midrule
    \vspace{-1.5mm}
    \\
    \toprule
    Black-box Data & \#Patient & \# Avg. Tokens & \# Train/Val/Test\\
    \midrule
    \textit{Operational Outcome} & 89 & 4,609 & - / - / 100 \\
    \textit{Lab Tests} & 100 & 3,907 & - / - / 100 \\
    \textit{New Diagnose} & 99 & 4,417 & - / - / 100 \\
    \bottomrule
\end{tabular}
}
\vspace{1mm}
\caption{Data Statistics. \textit{Operational Outcome}, include Long Length of Stay and ICU Transfer. \textit{Lab Tests} include Thrombocytopenia, Hyperkalemia, Hypoglycemia, Hyponatremia and Anemia. \textit{New Diagnose} include Hypertension, Hyperlipidemia, and Acute MI. \textbf{Top}: Clinical medical codes in structured, longitudinal format for assessing white-box model uncertainties. \textbf{Bottom}: Natural languages converted from medical sequence, for assessing uncertainties of proprietary GPTs.}
\label{tab:dataset_info}
\end{table}


\subsection{White-box Model Results}
\label{exp:white-box}
This section presents the Uncertainty Quantification findings of using BERT-based language models for EHR tasks. We first generate sequential embeddings from structured, sequential medical codes using CLMBR-T-base. We then follow the setting of~\citep{wornow2023ehrshot} to extract the medical codes' representation at prediction time point for downstream tasks. The prediction time point for General Operation Outcome tasks is at 11:59 pm on the day of admission and visits that last less than one day. For lab testing tasks, the prediction time point corresponds to one minute before the latest lab results are available. For New Diagnosis tasks, the prediction time is set to one minute before midnight on the day of discharge. We then trained a 2-layer Neural Network as decoders on the extracted representation embeddings for downstream tasks and reported the uncertainty metrics in the \textbf{top left} section of~\cref{tab:white-box}. We then implement Deep Ensemble and MC Dropout to reduce model uncertainties, results presented in the \textbf{top middle} and \textbf{top right} sections of ~\cref{tab:white-box}, respectively. For Deep Ensemble, we set the number of ensembling models to 5. For MC Dropout, we set the dropout ratio to 0.5. Second, we implement a multi-tasking approach such that a decoder can give predictions for clinical tasks within the same category. We begin by generating embedding for each task and combining it with the representations. Again, we implement the Deep Ensembles and MC Dropout for the multi-tasking pipeline with the same hyperparameter. We present the multi-tasking uncertainty metrics for the at the \textbf{bottom} of Table~\ref{tab:white-box}.

\vspace{-2mm}
\paragraph{Observations}
For both single-task settings and multi-task settings, Deep Ensemble consistently shows lower uncertainty metrics (measured by Brier score, NLL, ECE, and aECE) across most EHR tasks compared to the baseline. For new diagnosis tasks like "Thrombocytopenia" and "Hyperkalemia," Deep Ensemble improves considerably upon the baseline. The MC Dropout method also improves over the baseline but is generally less effective than the Deep Ensemble in reducing uncertainty. This indicates the effectiveness of the ensembling method in reducing the uncertainty of the model for clinical predictions, which prompted us to propose our initial methods to reduce the uncertainties of proprietary GPTs in~\cref{sec:reduce_uq}. In addition, the multi-task Deep Ensemble configuration consistently shows lower uncertainty metrics than the single-task configuration. Similar to Deep Ensemble, MC Dropout benefits from a multi-task setting, albeit with smaller margins of improvement. This indicates the benefits of our proposed multi-tasking method in EHR predictions, where tasks are within the same clinical category.

\begin{table*}[t]
\centering
    \resizebox{1
    \columnwidth}{!}{
    \begin{tabular}{lcc|cc|cc|cc|cc|cc}
    \toprule
    & \multicolumn{2}{c|}{\textbf{GPT-3.5 Single}} &\multicolumn{2}{c|}{\textbf{GPT-3.5 Multi}} 
    & \multicolumn{2}{c|}{\textbf{GPT-4 Single}} &\multicolumn{2}{c|}{\textbf{GPT-4 Multi}} 
    &\multicolumn{2}{c|}{\textbf{Ensemble Single}} &\multicolumn{2}{c}{\textbf{Ensemble Multi}}\\  
    \midrule
    & $Auc.$$\uparrow$ & $U.Q.$$\uparrow$ & $Auc.$$\uparrow$
    & $U.Q.$$\uparrow$ & $Auc.$$\uparrow$ & $U.Q.$$\uparrow$ 
    & $Auc.$$\uparrow$ & $U.Q.$$\uparrow$ & $Auc.$$\uparrow$
    & $U.Q.$$\uparrow$ & $Auc.$$\uparrow$ & $U.Q.$$\uparrow$ 
    \\
    Long Length of Stay & 0.5430 & 0.4570 & 0.6153 & 0.4265 & 0.3614 & 0.4992 & 0.5125 & 0.4875 & 0.5461 & \cellcolor{purple!6}0.8385 & \cellcolor{purple!6}0.6166 & 0.7237 \\
    ICU Transfer & 0.5047 & 0.5331 & 0.6853 & 0.3596 & 0.5938 & 0.5140 & 0.7083 & 0.4888 & 0.5538 & \cellcolor{purple!6}0.6455 & \cellcolor{purple!6}0.7552 & 0.4831 \\
    \midrule
    \vspace{-3.5mm}
    \\
    Thrombocytopenia & \cellcolor{purple!6}0.5327 & 0.4460 & 0.3745 & 0.5464 & 0.2917 & 0.5548 & 0.2062 & 0.5246 & 0.3235 & \cellcolor{purple!6}0.6382 & 0.2973 & 0.5594 \\
    Hyperkalemia & \cellcolor{purple!6}0.4795 & 0.4821 & 0.3673 & 0.5988 & 0.2508 & 0.5094 & 0.4020 & 0.5125 & 0.2553 & \cellcolor{purple!6}0.8470 & 0.3446 & 0.5948 \\
    Hypoglycemia & 0.5404 & 0.4410 & 0.5416 & 0.4352 & \cellcolor{purple!6}0.7131 & 0.4388 & 0.6688 & 0.4481 & 0.7052 & 0.5274 & 0.6386 & \cellcolor{purple!6}0.5723 \\
    Hyponatremia & \cellcolor{purple!6}0.4593 & 0.5220 & 0.3189 & 0.6303 & 0.2652 & 0.5347 & 0.2844 & 0.4939 & 0.2745 & 0.6186 & 0.2437 & \cellcolor{purple!6}0.6786 \\
    Anemia & \cellcolor{purple!6}0.4433 & 0.6100 & 0.2739 & 0.7254 & 0.1962 & 0.6877 & 0.2173 & 0.6044 & 0.2037 & \cellcolor{purple!6}0.7722 & 0.2069 & 0.6825 \\
    \midrule
    \vspace{-3.5mm}
    \\
    Hypertension & 0.4762 & 0.5238 & 0.5758 & 0.4707 & \cellcolor{purple!6}0.7136 & 0.4488 & \cellcolor{purple!6}0.7136 & 0.4222 & 0.6920 & 0.5606 & 0.7042 & \cellcolor{purple!6}0.6804 \\
    Hyperlipidemia & 0.5559 & 0.4203 & 0.5478 & 0.5032 & 0.6289 & 0.4077 & 0.6845 & 0.4127 & \cellcolor{purple!6}0.7109 & 0.4014 & 0.6736 & \cellcolor{purple!6}0.7338 \\
    Acute MI & 0.5455 & 0.4545 & 0.4929 & 0.6379 & \cellcolor{purple!6}0.7317 & 0.3394 & 0.6149 & 0.3436 & 0.6403 & 0.6053 & 0.6074 & \cellcolor{purple!6}0.7335 \\
    \bottomrule
    \end{tabular}
    }
    \caption{\small
    Uncertainty Metrics for Clinical Predictions with Proprietary GPTs. \textbf{Single} refers to the setting where answers are generated for one task at a time. \textbf{Multi} refers to the setting where multiple answers are generated simultaneously for tasks within the same clinical category. \textbf{Ensemble} refers to the approach that combines responses from multiple proprietary GPTs.
    }
    \vspace{-5mm}
    \label{tab:gpt}
\end{table*}

\subsection{Black-box GPT Results}
\label{exp:black-box}

We conduct evaluations of our black-box uncertainty quantification methods using the specifically tailored EHRSHOT dataset, as detailed in~\cref{tab:dataset_info}. For each task category, we began by filtering the EHRSHOT dataset based on prediction time points; we ensured that each instance of converted medical language had corresponding ground truth labels for all tasks within the same category. We then constructed the test set for each task category by stochastically sampling 100 entries from the tailored EHRSHOT dataset. Our sampling algorithm was designed to terminate once the data for all subtasks contained at least 12 positive labels. For evaluations, we employ GPT-4 and GPT-3.5 Turbo to generate responses. We repeatedly generate five responses for each constructed prompt. We then calculate the $AUC$ score by cleaning the outputs and matching the generated answers with the ground-truth clinical outcomes. We then calculate the uncertainty metric $U.Q.$ in~\cref{eq:ce} and use it to predict whether the response from GPTs is correct. Similar to the multitasking method described in~\cref{exp:white-box}, we reformulated our prompts to request predictions for multiple clinical tasks within the same category. Furthermore, akin to the Deep Ensemble approach in the white-box setting, we aggregate the responses from multiple GPTs and calculate the $AUC$s and the uncertainty metrics. Our results are presented in Table~\ref{tab:gpt}.

\vspace{-2mm}
\paragraph{Observations} In evaluating model performances, GPTs show limited performance in predicting clinical outcomes from free-formed languages. In evaluation uncertainties, for GPT-3.5-Turbo and GPT-4, the ensembling methods score higher in UQ metrics in almost every case listed than in the single-model setting. In addition, GPT-4 ensembles outperform single-model significantly in UQ metrics. This indicates that ensembling models perform significantly better in quantifying thus reducing uncertainty across all clinical prediction tasks than their single counterparts. We observe minimal or no improvements in analyzing the UQ metric between single-tasking and multi-tasking settings within individual GPT models. Tasks such as Hypokalemia and Hyponatremia exhibit similar U.Q. scores regardless of whether they are approached through single or multi-task configurations within the same models. However, when multi-tasking is integrated with ensemble methods, we observe a marginal improvement in U.Q metrics. This indicates that combining multi-tasking approaches within ensemble frameworks substantially contributes to more reliable assessments of whether to trust LLM's output, thus reduce the uncertainties for EHR predictions.

\vspace{-2mm}
\section{Conclusion}
\label{sec:conclusion}
\vspace{-2mm}

In this work, we explored the quantification and reduction of uncertainty in clinical outcome predictions with EHR by harnessing white-box language models and black-box large language models. We focused on two main methodologies to mitigate uncertainty: ensemble methods, which combine predictions from multiple models, and multi-tasking, where models simultaneously predict multiple clinical outcomes. By transferring and adapting these methodologies originally developed for LMs to the realm of LLMs, we demonstrated reductions in uncertainties across white-box and black-box models. 

\vspace{-2mm}
\section{Limitations}
Our methods were validated using clinical prediction tasks containing longitudinal EHRs. While the results were promising, the generalizability of these methods to other distinct domains with limited data has not yet been tested. Future work may explore data from different cultures, and the adaptation of these methods for broader applications beyond the current scope.

\section*{Acknowledgments}
This work was completed by the first author during their internship at GE Healthcare.

\bibliographystyle{unsrt}  
\bibliography{references}

\end{document}